\documentclass[nonacm=true,sigconf,authorversion]{acmart} 

\usepackage[UTF8, scheme=plain, punct=plain, zihao=false]{ctex}
\usepackage{algorithm}
\usepackage{algpseudocode}
\usepackage{subcaption}
\usepackage{multirow}
\usepackage{makecell}

\AtBeginDocument{%
  }

\setcopyright{acmlicensed}
\copyrightyear{2026}
\acmYear{2026}
\acmDOI{XXXXXXX.XXXXXXX}
\acmConference[Conference acronym 'XX]{Make sure to enter the correct
  conference title from your rights confirmation email}{June 03--05,
  2026}{Woodstock, NY}
\acmISBN{978-1-4503-XXXX-X/2018/06}

\begin{document}

\title{HART: Data-Driven Hallucination Attribution and Evidence-Based Tracing for Large Language Models}

\author{Shize Liang}
\affiliation{%
	\institution{Faculty of Computing, Harbin Institute of Technology}
	\city{Harbin}
	\country{China}}
\email{2023111781@stu.hit.edu.cn}

\author{Hongzhi Wang}
\authornote{Corresponding author.} 
\affiliation{%
	\institution{Faculty of Computing, Harbin Institute of Technology}
	\city{Harbin}
	\country{China}}
\email{wangzh@hit.edu.cn}

\renewcommand{\shortauthors}{Trovato et al.}

\begin{abstract}
  Large language models (LLMs) have demonstrated remarkable performance in text generation and knowledge-intensive question answering. Nevertheless, they are prone to producing hallucinated content, which severely undermines their reliability in high-stakes application domains.
  Existing hallucination attribution approaches, based on either external knowledge retrieval or internal model mechanisms, primarily focus on semantic similarity matching or representation-level discrimination. As a result, they have difficulty establishing structured correspondences at the span level between hallucination types, underlying error generation mechanisms, and external factual evidence, thereby limiting the interpretability of hallucinated fragments and the traceability of supporting or opposing evidence.
  To address these limitations, we propose \textit{HART}, a fine-grained hallucination attribution and evidence retrieval framework for large language models. \textit{HART} formalizes hallucination tracing as a structured modeling task comprising four stages: \emph{span localization, mechanism attribution, evidence retrieval, and causal tracing}. Based upon this formulation, we develop the first structured dataset tailored for hallucination tracing, in which hallucination types, error mechanisms, and sets of counterfactual evidence are jointly annotated to enable causal-level interpretability evaluation. 
  Experimental results on the proposed dataset demonstrate that \textit{HART} substantially outperforms strong retrieval baselines, including \textit{BM25} and \textit{DPR}, validating the effectiveness and generalization capability of the proposed tracing paradigm for hallucination analysis and evidence alignment.
\end{abstract}



\received{20 February 2007}
\received[revised]{12 March 2009}
\received[accepted]{5 June 2009}

\maketitle

\section{Introduction}
	Large language models (LLMs) demonstrate outstanding abilities in text generation, knowledge question-answering, reasoning, and summarization tasks, finding widespread application in high-stakes domains such as healthcare, finance, and law~\cite{matarazzo2025surveylargelanguagemodels,Wang2023ClinicalGPT}. Even so, regardless of their superior performance in natural language understanding and generation, LLMs frequently exhibit "hallucination". This refers to generating statements that appear plausible on the surface, but are factually incorrect or lack reliable evidence. In high-stakes domains, such hallucinations can lead to severe consequences, including misleading medical assessments, incorrect legal references, and the spread of false information~\cite{Han2024,10.1093/jla/laae003, Tang2023MedicalSummarization}. It means that developing accurate and efficient attribution mechanisms to locate, analyze, and trace the origins of hallucinated content has become a critical research direction for enhancing the reliability and interpretability of LLMs.
	
	\begin{figure*}[h]
		\centering
		\includegraphics[width=\linewidth]{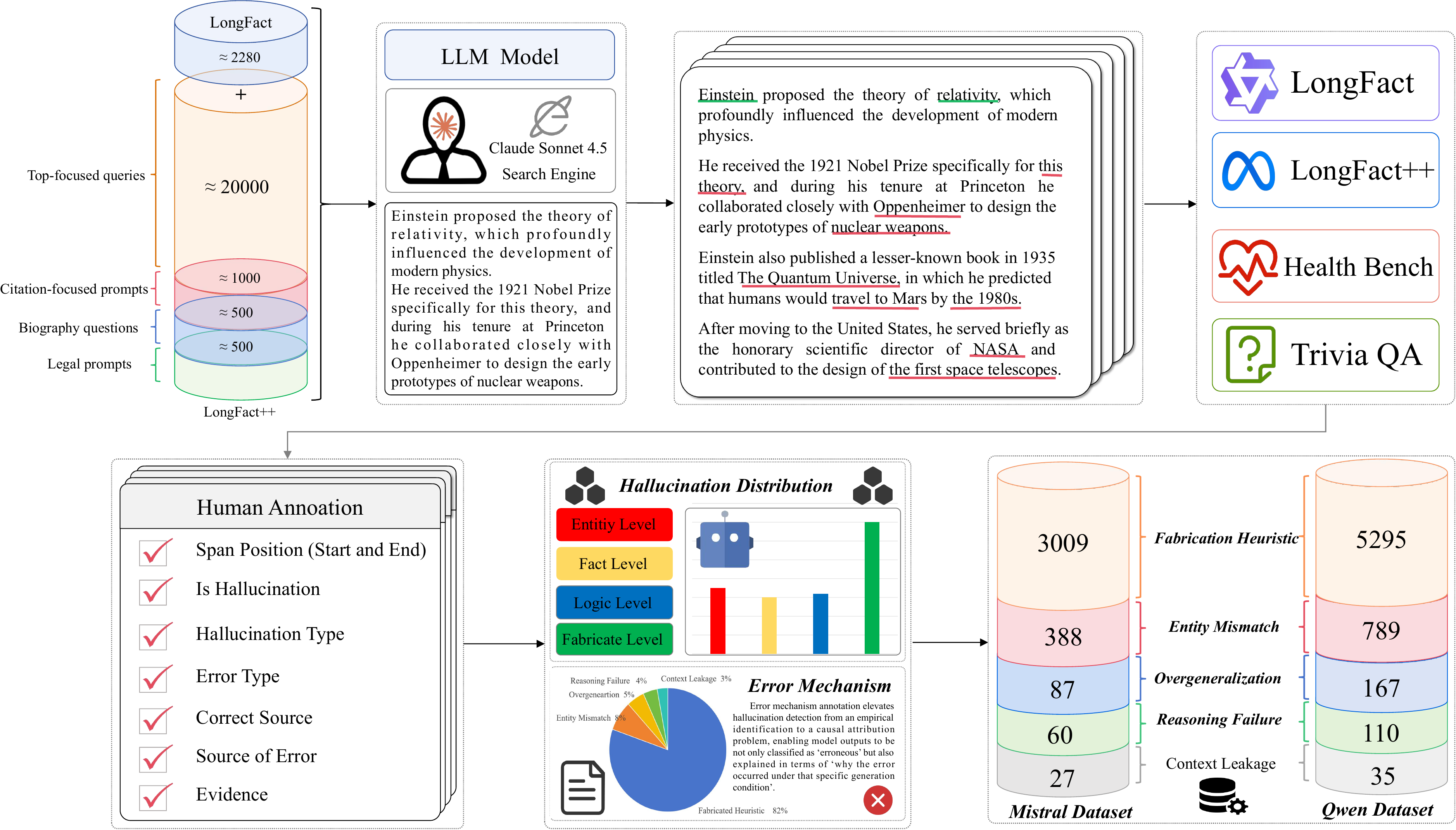}
		\caption{The construction of the hallucination tracing dataset employs a labeling framework combining large language model assistance with human supervision. }
		\label{fig:dataset}		
	\end{figure*}
	
	In recent years, academia has conducted extensive exploration on the issue of hallucinations. A particular area of research focuses on the field of hallucination detection, relying on methods such as uncertainty estimation, external knowledge retrieval, and model internal states to identify potential hallucinated content in model outputs~\cite{Farquhar2024,zhang-etal-2023-enhancing-uncertainty,yehuda-etal-2024-interrogatellm, su2024unsupervisedrealtimehallucinationdetection}. While detection performance has improved, most studies remain at the superficial level of "error detection", failing to distinguish between hallucination types and error mechanisms at a granular level. This hinders structured analysis of the causes of hallucinations.
	The other mainstream paradigm focuses on investigating the origins and generation mechanisms of hallucinations. Through examining dimensions such as internal representations, generation strategies, and semantic distributions within models, it aims to reveal the intrinsic patterns and driving factors behind hallucination production~\cite{ijcai2025p929,ji-etal-2024-llm}. Yet, such approaches remain confined to explanations of macroscopic interpretability. They lack a closed-loop evaluation framework integrating semantic retrieval with external factual evidence verification, making it difficult to directly pinpoint specific erroneous segments or align them with external evidence.

    Both paradigms exhibit significant limitations in research objectives and methodology. The former focuses on whether hallucinations exist, and even when locating errors at the fragment level, its analysis remains confined to the identification stage, lacking a systematic description of hallucination types and error generation mechanisms. The latter concentrates on explaining why hallucinations occur through internal model mechanisms, yet related analyses are largely restricted to latent model representations. Neither approach establishes a systematic mapping to objective, verifiable, and alignable facts in the real world.
	
	Thus, existing research faces a common challenge at the mechanism level: hallucinations fundamentally stem from the model's gradual deviation from local factual assertions in the generation process. However, current studies lack both a unified modeling of fine-grained hallucination types and error generation mechanisms, as well as explicit alignment and attribution mechanisms linking hallucination fragments to external objective facts. This makes it difficult to answer critical questions about hallucination fragments: "At the factual level, where exactly did the error occur?" and "What is the actual fact?"
	
    To answer these questions, we attempt to explicitly trace each hallucination fragment back to its underlying error mechanism and corresponding external factual evidence through a unified attribution–retrieval framework, thereby uncovering the sources of hallucinations and establishing causal relationships with external factors. Based on this idea, we propose \textit{HART (Hallucination Attribution Retrieval Tracing)}, a framework for hallucination attribution and evidence retrieval tracing tailored for large language models. By integrating a data-driven classification system with semantic retrieval mechanisms, it constructs causal chains that link model outputs to external knowledge evidence. This allows for systematic characterization and interpretable analysis of hallucination origins and generation mechanisms.
	
	In summary, the main contributions of this paper are summarized as follows:

    \textbullet \ \ We are pioneers in formulating hallucination analysis from the perspective of external factual evidence tracing, modeling hallucination as an attribution and evidence-tracking task grounded in the objective world. This reframes hallucination research from a paradigm centered on internal mechanism analysis and output-level detection toward a causal tracing framework driven by real-world fact coherence.

    \textbullet \ \ We propose \textit{HART} framework, which unifies hallucinated span classification, error mechanism attribution, and external evidence alignment within a single paradigm, establishing a complete causal tracing pipeline from model-generated content to verifiable factual grounds.

    \textbullet \ \ We construct the first fine-grained, structured hallucination tracing dataset at the span level, providing multi-dimensional annotations over hallucination types, error mechanisms, and evidence sets to support causal-level interpretability evaluation. On this dataset, we systematically demonstrate the significant performance advantages of \textit{HART} over existing baseline methods.

\vspace{-1em}
\section{Related Work}
	
	\subsection{RAG}
	Retrieval-Augmented Generation (RAG) is recognized as one of the core paradigms used in contemporary hallucination detection and factual consistency verification. Initial studies introduced external knowledge bases into the generation process, establishing explicit links between model outputs and external evidence~\cite{NEURIPS2020_6b493230}, thereby establishing RAG technology as the foundational framework for hallucination detection research. Building upon this foundation, RAGTruth constructs a span-level hallucination annotation dataset tailored for RAG scenarios, providing fine-grained evaluation criteria for unsupported statements, evidence conflicts, and fabricated information within generated content~\cite{niu-etal-2024-ragtruth}.
	Subsequent work shifted focus to how models genuinely utilize retrieved evidence. LUMINA and LRP4RAG introduced contextual knowledge signals and Layer-wise Relevance Propagation mechanisms, respectively, to characterize the dependency between generated content and retrieved documents from an internal model perspective, further enhancing the robustness and interpretability of hallucination detection methods across domains and noisy scenarios~\cite{liu2025unsupervisedmonoculardepthestimation,hu2025lrp4ragdetectinghallucinationsretrievalaugmented}.
	
	\subsection{Model internal mechanism traceability}
	The internal mechanism tracing of models originates from the computational process itself, attempting to clarify the generation pathways of hallucinations within the internal semantic space. MIND trains lightweight detectors based on LLM hidden states, demonstrating that hallucinations possess separable structural features within the internal representation space. This advances hallucination detection from external empirical validation toward internal signal modeling~\cite{su-etal-2024-unsupervised}. Developing upon this basis, researchers systematically investigate the semantic drift phenomenon in hidden states and attention distributions during decoding by introducing contextual perturbations and attention tracking~\cite{wei2025shadowsattentioncontextualperturbation}.
	To further enhance interpretability depth, researchers adopted a causal perspective to systematically model internal reasoning structures. The DST method constructs semantic causal graphs between hidden layers, mapping dependency relationships and conflict patterns within computational pathways to clarify when and where patterns deviate from reality within internal inference routes~\cite{bhatia2025distributionalsemanticstracingframework}. Meanwhile, such approaches remain confined to the model's internal representation space, struggling to provide verifiable, traceable external evidence supporting specific hallucination fragments.
	
	As internal mechanism modeling frameworks grow in sophistication, certain studies concentrate on the stability of internal semantic structures in out-of-distribution scenarios. By designing guiding prompts, these approaches aim to make the model's internal state more explicitly encode truth signals, which improves generalization across domains. However, this method remains confined to the internal discriminative level, struggling to provide traceable evidence corresponding to real-world knowledge for specific erroneous fragments~\cite{zhang-etal-2025-prompt}. To deal with this, \textit{HART} introduces an evidence repository retrieval and semantic alignment mechanism. It maps model-generated hallucinations to verifiable real-world sources, achieving a paradigm shift from "internal separability discrimination" to "externally verifiable evidence attribution".

\section{Methodlogy}

    \subsection{Overview}
   	The proposed \textit{HART} method follows a unified framework of "data-driven modeling $\rightarrow$ evidence retrieval $\rightarrow$ causal attribution", systematically modeling the hallucination attribution problem as a text-fragment-oriented process of attribution and fact alignment.
    
    Specifically, we first construct a fine-grained, structured dataset for hallucination attribution by systematically annotating hallucinated segments in model-generated text, providing a supervised foundation for subsequent causal analysis (Section~\ref{sec:dataset}).
    To retrieve relevant information, we design a two-stage semantic retrieval framework that efficiently identifies external evidence semantically aligned and factually contrastive with hallucinated spans within a large-scale evidence space (Section~\ref{sec:retrieval}).
    Finally, we integrate hallucination type classification, error mechanism attribution, and evidence retrieval results into a unified modeling paradigm to produce span-level hallucination attribution outputs, enabling systematic characterization and interpretable analysis of hallucination causes and their factual grounding (Section~\ref{sec:trace}).
 
	\subsection{Datasets Construction}
    \label{sec:dataset}
	
	\begin{figure*}[h]
		\centering
		\includegraphics[width=\linewidth]{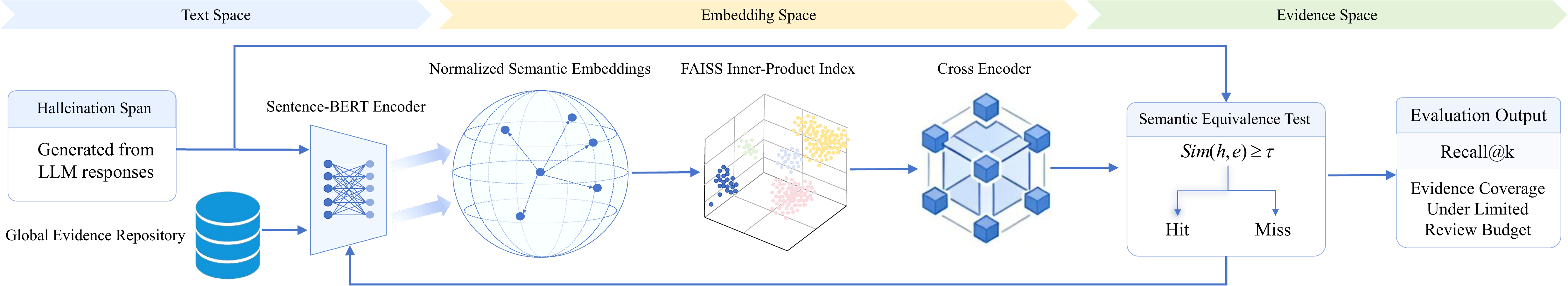}
		\caption{Evidence Retrieval Process: Leveraging semantic embeddings, vector-based indexing, and a Cross-Encoder fine-ranking mechanism, this process aligns hallucinated fragments with external factual evidence and retrieves the most relevant supporting information.}
		\label{fig:retrieve}
	\end{figure*}
	
	 \figurename ~\ref{fig:dataset} illustrates the annotation process for the hallucination attribution dataset. This paper models hallucination attribution as a fine-grained structured annotation problem oriented toward "text fragment $\rightarrow$ hallucination category classification $\rightarrow$ error mechanism attribution $\rightarrow$ evidence traceability", thereby enabling interpretable and causally hierarchical evaluation of evidence for model hallucination attribution.
	
	The formal definition of the hallucination attribution dataset constructed in this paper is as follows:
	\begin{equation}
		\mathcal{D} = \{(q_{i}, r_{i}, \mathcal{H}_{i})\}_{i=1}^{N}, \quad
		h_{ij} = (s_{ij}, z_{ij}, y_{ij}, m_{ij}, a_{ij}, \mathcal{E}_{ij})
	\end{equation}
	Let $q_{i}$ denote the user query prompt, $r_{i}$ the model-generated text, and $\mathcal{H}_{i}$ the set of hallucination fragments specified in character span format. For each fragment $h_{ij}$, the dataset further provides multidimensional structured annotations, including the hallucination presence label $z_{ij}$, hallucination type $y_{ij}$, error mechanism $m_{ij}$, error attribution source $a_{ij}$, and the set of adversarial evidence $\mathcal{E}_{ij}$ semantically aligned with the fragment.
	
	We utilize the long-text entity dataset \textit{LongFact++} as the source of factual semantic constraints. This dataset expands upon the factual \textit{LongFact} dataset and utilizes the \textit{Claude Sonnet 4.5} model with web search capabilities to annotate hallucinated fragments. The above procedure maximizes $max \mathbb{E}_{x \sim Q}[\Delta(r(x), C)|z = 1]$ as the objective function, enhancing semantic deviation diversity while preserving the hallucination distribution of the real-world model.
		
	For recognized hallucination segments, this paper proposes a dual-layer annotation framework. The first layer reflects the surface manifestations of hallucinations, including entity hallucinations, factual hallucinations, logical hallucinations, and fabricated hallucinations. The second layer characterizes the underlying error mechanisms of hallucinations, encompassing five types: entity mismatch, overgeneralization, reasoning failure, context leakage, and fabrication heuristic.

	Regarding each hallucination segment $s_{ij}$, we construct an evidence base $\mathcal{B}$ using \textit{Wikipedia} and authoritative official websites, from which the \textit{ChatGPT 5.1} model equipped with web retrieval capabilities is employed to perform optimal evidence selection within the candidate evidence space, enabling the construction of an adversarial evidence set that achieves high semantic alignment with hallucination segments while minimizing redundancy, formalized as follows:
	\begin{equation}
		\mathcal{E}^{*} = \arg\max_{\hat{\mathcal{E}} \subset \mathcal{B}} \sum_{e \subset \hat{\mathcal{E}}} \text{Rel}(e, s) - \lambda \cdot \text{Red($\hat{\mathcal{E}}$)}
	\end{equation}
	In this case, $\text{Rel}(\cdot)$ indicates the semantic relevance between evidence and hallucination fragments, while $\text{Red}(\cdot)$ represents the evidence redundancy penalty term.
	
	The annotation process utilizes a noise control mechanism combining "automatic annotation with human-supervised refinement". The initial annotation $A_{0}$ is generated by LLM, while the final structured annotation $A$ is derived from a human-refinement function. During this process, we adopt a batch sampling strategy with a fixed window size $w$. The annotation noise rate is defined as:
	\begin{equation}
		\epsilon_{w} = \frac{1}{w} \sum_{i=1}^{w} \mathbb{I}(A_{i}^{LLM} \neq A_{i}^{human})
	\end{equation}
	And constrain $\epsilon_{w} \le \tau$, otherwise trigger the re-labeling process.
		
	The final constructed dataset consists of quadruplets in the form of $\{s, y, m, \mathcal{E}\}$. This structure enables collaborative evaluation of precise localization of model outputs, comprehension of hallucination types and error mechanisms, and verification of evidence consistency.
		
	\subsection{Retrieval}
    \label{sec:retrieval}
	We propose an evidence retrieval framework based on semantic representation learning and vector-based nearest neighbor search, built upon the causal-level hallucination detection dataset. The framework treats hallucinated fragments as query units and aligns them with counter-evidence in a high-dimensional semantic space, enabling automatic tracing of model-generated errors and evaluation of evidence-level correction capabilities.
	
	\figurename ~\ref{fig:retrieve} illustrates the overall workflow of this framework, encompassing four stages: evidence document construction, semantic embedding encoding, vector indexing, and proximity retrieval with hit evaluation. This forms a closed-loop validation mechanism that transitions from the textual space to the semantic space and back to the evidence textual space.
		
	\subsubsection{Evidence-level Document Corpus Construction}
	In the hallucination traceability dataset, each hallucination segment $s_{ij}$ is accompanied by corresponding error correction sources and adversarial evidence texts. Therefore, we uniformly extract the evidence sets $\mathcal{E}_{ij}$ from the samples and organize them into a consolidated evidence document corpus $\mathcal{E}$. Each piece of the evidence is indexed and retrieved as an independent semantic unit, elevating the evidence structure originally attached to the sample level to a globally shared semantic retrieval space. This design facilitates evidence reuse and enables consistency evaluation across different samples and contexts.
	
	\subsubsection{Semantic Vector Encoding}
	To support comparison between hallucinated fragments and evidence texts within a unified semantic space, we employ the Sentence-BERT model to vectorize text, mapping natural language onto a shared embedding space. Within this space, we assume the following properties:
		
	\textbullet \ \ Semantic proximity preservation: Semantically consistent or factually related texts exhibit smaller vector distances in the embedding space.
		
	\textbullet \ \ Semantic separability: Text with conflicting facts or semantic inconsistencies exhibits significant spatial separation.
		
	Based on these assumptions, the evidence retrieval problem can be formally defined as: identifying the set of proximate evidence within the embedding space that is most semantically relevant to a given hallucination fragment while holding opposing factual positions.
		
	Specifically, for any evidence text $e_{j}$, it is first mapped to a vector representation $v_{j}$ via the encoding function $f(\cdot)$, followed by $L_{2}$ normalization processing:
	\begin{equation}
		v_{j} = f(e_{j}), \quad
		\hat{v_{j}} = \frac{v_{j}}{\|v_{j}\|_{2}}
	\end{equation}
		
	After normalization, the vector dot product is equivalent to cosine similarity:
	\begin{equation}
		\langle \hat{v_{a}}, \hat{v_{b}} \rangle = \cos(v_{a}, v_{b})
	\end{equation}
	This procedure renders the retrieval process geometrically identical to performing maximum angular similarity searches on a unit hypersphere, ensuring consistent scaling in similarity calculations across texts of varying lengths.

	\subsubsection{Vector Indexing and Retrieval}
	In order to achieve efficient retrieval over large-scale evidence repositories, we construct a FAISS-based vector indexing structure using inner-product similarity as the matching metric. This ensures high-precision semantic matching while reducing the computational complexity of evidence retrieval from linear to sublinear time. During retrieval, each hallucination fragment is encoded as a query vector. The system returns the Top-$k$ evidence candidates with the highest semantic similarity in the index space. This procedure is theoretically equivalent to finding an approximate solution to the following optimization problem:
	\begin{equation}
		e^{*} = \arg\max_{e \in \mathcal{E}} Sim(s_{ij}, e)
	\end{equation}
		
	Here, Sim($\cdot$) denotes a semantic similarity metric in the embedded space.
		
	This proximity search mechanism implicitly assumes that if authentic evidence contradicting the hallucinated fragment exists, it should maintain high relevance to the hallucination within its semantic space, thereby being retrieved with high probability in the Top-$k$ set.
	
	\begin{figure*}[h]
		\centering
		\includegraphics[width=\linewidth]{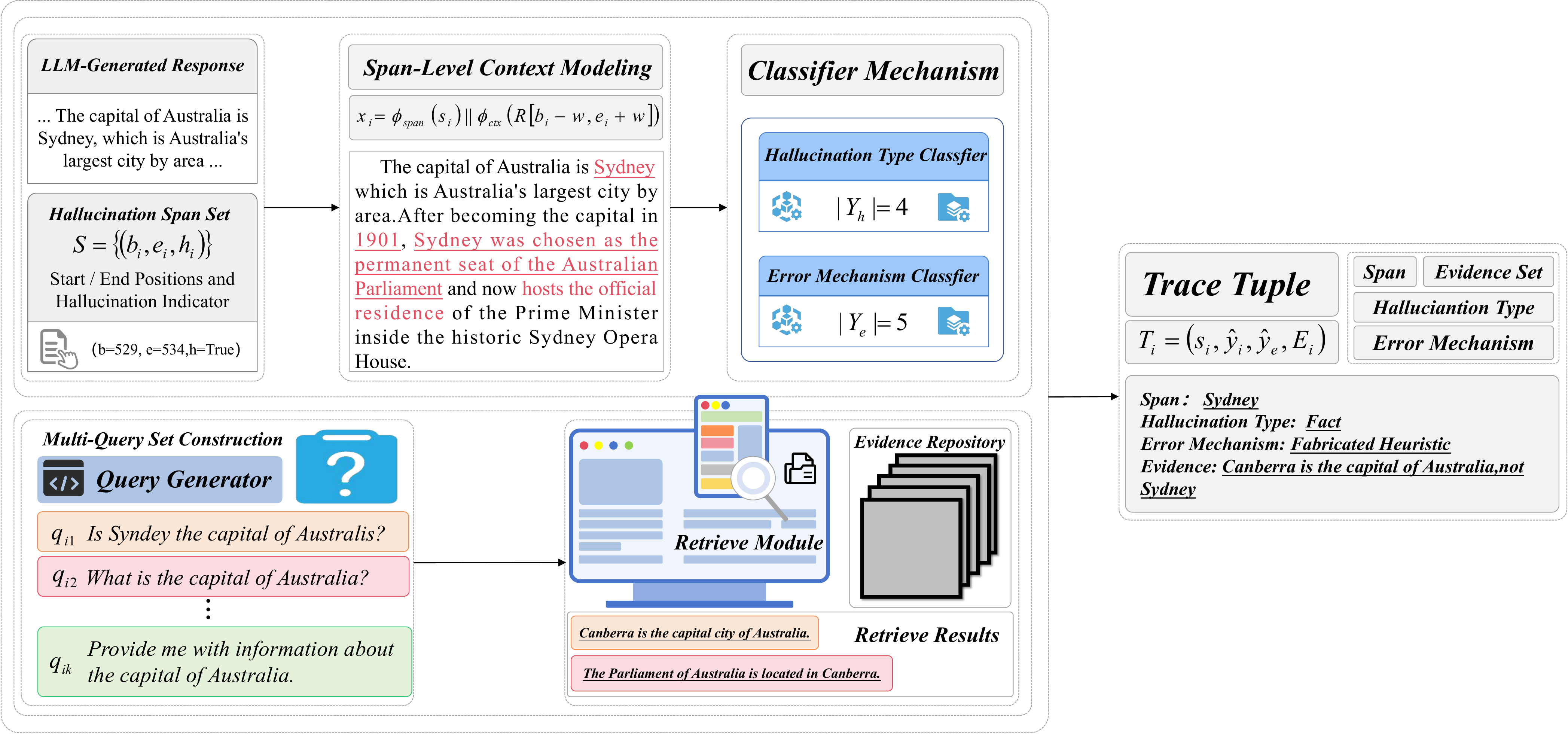}
		\caption{Span-level hallucination tracing workflow: \textit{HART} integrates hallucination type classification, error mechanism attribution, and evidence retrieval to systematically model and trace the external evidence supporting each hallucinated segment in model-generated text.}
		\label{fig:trace}
	\end{figure*}
		
	\subsubsection{Evidence Reranking and Semantic Refinement}
	
	The Top-$k$ candidate evidence set returned by vector indexing ensures high recall yet relies solely on geometric proximity in the embedding space for similarity measurement, making it difficult to fully capture fine-grained semantic alignment and factual consistency. To tackle this, this paper introduces a Cross-Encoder-based fine-ranking model after the coarse retrieval phase to perform a two-stage re-scoring of deep semantic relationships between queries and evidence.
	
	In particular, for each hallucinated segment $s_i$ and its query set $\mathcal{Q}_i = \{q_{ij}\}$, we jointly encode all query-document pairs $(q_{ij}, d)$ from the coarse retrieval candidate evidence set $\mathcal{D}_{ij}$ and compute a refined semantic matching score:
	\begin{equation}
		s(q_{ij}, d) = \text{CrossEncoder}(q_{ij}, d)
	\end{equation}
	
	This scoring function explicitly models cross-sentence alignment relationships and factual consistency constraints between queries and evidence texts, elevating evidence ranking from "geometric similarity" to "semantically interpretable" outcomes.
	
	In accordance with this, for each hallucinated fragment $s_i$, we select the evidence set with the highest precision-ranking score from all candidate sets corresponding to the query as the final attribution output:
	\begin{equation}
		\mathcal{E}_i = \arg\max_{j} \max_{d \in \mathcal{D}_{ij}} s(q_{ij}, d)
	\end{equation}	
	
	\subsubsection{Hit Determination and Theoretical Consistency}
	To mitigate evaluation bias induced by superficial string matching, this paper adopts a "semantic similarity + contextual mechanism" criterion for determining retrieval hits. Specifically, query spans are constructed using multiple contextual window sizes. A hallucinated fragment is considered successfully retrieved if any evidence in the retrieval results reaches a predefined semantic similarity threshold with the manually annotated evidence in the embedding space.
	
	This evaluation strategy can be interpreted as a semantic equivalence class determination mechanism: it does not require the system to retrieve identical evidence texts but permits retrieval of alternative evidence that is semantically equivalent at the factual level or supports the same factual proposition. Such a design reflects the model's retrieval and reasoning capabilities in real-world environments.
		
	Retrieval performance is measured using $\text{Recall@}k$. Owing to space constraints, the rationale for selecting this metric is provided in the appendix~\ref{appendix_a}.

	\subsection{Evidence Tracing}
    \label{sec:trace}
	
	\begin{algorithm}[h]
		\caption{Span-Level Hallucination Tracing Framework}
		\label{alg:trace}
		\begin{algorithmic}[1]
			\State \textbf{Input:} Model-generated text $R$, hallucination span set $\mathcal{S}$, semantic encoder $f(\cdot)$, evidence corpus $\mathcal{B}$, FAISS vector index $\mathcal{I}$, Cross-Encoder reranker $\mathcal{C}$, context window size $w$, and number of queries $k$
			\State \textbf{Output:} Hallucination tracing results $\{\mathcal{T}_i\}$
			
			\Function{Trace}{$R, \mathcal{S}, f, \mathcal{B}, \mathcal{I}, \mathcal{C}, w, k$}
			
			\State Initialize the result set $\mathcal{T} \leftarrow \emptyset$
			
			\For{Each hallucination span $s_i = (b_i, e_i, h_i) \in \mathcal{S}$}
			
			\State $\hat{y}_i^h \leftarrow \arg\max_{y \in \mathcal{Y}_h} p(y \mid x_i)$
			
			\State $\hat{y}_i^e \leftarrow \arg\max_{y \in \mathcal{Y}_e} p(y \mid x_i)$
			
			\State $\mathcal{Q}_i \leftarrow \{q_{i1}, q_{i2}, \dots, q_{ik}\}$ 
			
			\State Initialize candidate evidence set $\mathcal{D}_i \leftarrow \emptyset$
			
			\For{Each query $q_{ij} \in \mathcal{Q}_i$}
			\State $\hat{q}_{ij} \leftarrow f(q_{ij})$
			\State $\mathcal{D}_{ij} \leftarrow \mathcal{I}(\hat{q}_{ij})$
			\State $\mathcal{D}_i \leftarrow \mathcal{D}_i \cup \mathcal{D}_{ij}$
			\EndFor
			
			\For{Each candidate evidence $d \in \mathcal{D}_i$}
			\State $s_i(d) \leftarrow \mathcal{C}(s_i, d)$
			\EndFor
			
			\State $\mathcal{E}_i \leftarrow \arg\max_{d \in \mathcal{D}_i} s_i(d)$
			
			\State $\mathcal{T}_i \leftarrow (s_i, \hat{y}_i^h, \hat{y}_i^e, \mathcal{E}_i)$
			\State $\mathcal{T} \leftarrow \mathcal{T} \cup \{\mathcal{T}_i\}$
			
			\EndFor
			
			\State \Return $\mathcal{T}$
			\EndFunction
		\end{algorithmic}
	\end{algorithm}
	
	\figurename ~\ref{fig:trace} showcases the fine-grained hallucination attribution framework \textit{HART} proposed in this paper. By treating hallucinated segments in model-generated text as fundamental units, it integrates modules for hallucination type identification, error mechanism attribution, and evidence retrieval to structurally model and systematically characterize the external evidence leading to each hallucination span.
	
	\subsubsection{Problem Definition}
	Given the generated text $R$ from a large language model, along with a manually annotated collection of hallucination fragments.
	\begin{equation}
		\mathcal{S} = \{s_{i} = (b_{i}, e_{i}, h_{i})\}
	\end{equation}
	
	Below, $b_i$ and $e_i$ symbolize the start and end positions of the hallucination segment within the generated text $R$, respectively, while $h_i \in \{0,1\}$ indicates whether the segment is labeled as a hallucination.
	
	For each identified hallucination segment $s_{i}$, this paper aims to perform the following joint prediction and retrieval tasks:
	
	\textbullet \ \ Determine the corresponding hallucination type.
	
	\textbullet \ \ Attribute the underlying error generation mechanism.
	
	\textbullet \ \ Retrieve the most likely evidence set from external corpora that supports or refutes the segment, thereby validating its factual stance.

	\subsubsection{Span-Level Context Modeling}
	
	To capture the local context and semantic dependencies of hallucinated fragments, we adopt a "\textit{Span + Context}" input construction strategy. Specifically, for a hallucinated fragment $s_{i}$, context windows of length $w$ are extracted from both its preceding and succeeding positions to form the model input representation:
	\begin{equation}
		\begin{aligned}
			x_{i} = \phi_{\text{span}}(s_{i}) ~ \Vert ~ \phi_{\text{ctx}}(R[b_i - w : e_i + w])
		\end{aligned}
	\end{equation}
	Here, $\phi_{\text{span}}(\cdot)$ and $\phi_{\text{ctx}}(\cdot)$ denote the span-marking function and the context-marking function, respectively. This representation aligns hallucination boundaries with the generation context, enabling simultaneous modeling of fragment-level semantic features and contextual dependency structures during both hallucination type discrimination and error mechanism identification, thereby improving robustness and generalization under complex contextual conditions.
	
	\subsubsection{Hallucination Type and Error Mechanism Classification}
	
	This paper models hallucinations from two complementary dimensions: surface manifestations and underlying generation mechanisms, corresponding to the two categories of labels: hallucination types and error mechanisms. To avoid semantic coupling and interference between label spaces, we employ two parameter-independent classification models to perform decoupled predictions on these attributes.
	
	Set up the hallucination type discrimination function: 
	\begin{equation}
		f_h : \mathcal{S} \rightarrow \mathcal{Y}_h, \quad |\mathcal{Y}_h| = 4
	\end{equation}
	
	Given $y_h \in \mathcal{Y}_h$, define the error mechanism attribution function:
	\begin{equation}
		f_e : \mathcal{S} \times \mathcal{Y}_h \rightarrow \mathcal{Y}_e, \quad |\mathcal{Y}_e| = 5
	\end{equation}
	
	For each input $x_{i}$, we construct separate hallucination-type classifiers and error mechanism classifiers based on the BERT architecture, and perform predictions using the maximum a posteriori (MAP) criterion:
	\begin{equation}
		\hat{y}_i^h = \arg\max_{y \in \mathcal{Y}_h} p(y \mid x_i), \quad
		\hat{y}_i^e = \arg\max_{y \in \mathcal{Y}_e} p(y \mid x_i)
	\end{equation}
	
	This decoupled modeling approach enables the system to distinguish between "hallucination types" and "error mechanism", thus offering structured supervisory signals for subsequent causal-level root-cause analysis.
	
	\begin{table*}[h]
		\centering
		\caption{Ablation Study on Evidence Retrieval Performance}
		\large
		\begin{tabular}{lcccccc}
			\toprule
			Method & R@1 & R@5 & R@10 & MRR & nDCG@5 & nDCG@10 \\
			\midrule
			Dense Embedding & 0.4133 & 0.6387 & 0.7146 & 0.5092 & 0.5340 &
			0.5586 \\
			Dense Embedding + Cross-Encoder & 0.5172 & 0.6868 & 0.7146 & 0.5887 & 0.6104 & 0.6196 \\
			Dense Embedding + Multi-Query & 0.6244 & 0.8058 & 0.8557 & 0.7016 & 0.7227 & 0.7389\\
			Ours (HART) & \textbf{0.7068} & \textbf{0.8360} & \textbf{0.8557} & \textbf{0.7619} & \textbf{0.7786} & \textbf{0.7850}\\
			\bottomrule
		\end{tabular}
		
		\label{table:ablation}
	\end{table*}
	
	\subsubsection{Evidence Retrieval and Optimal Trace Selection
	}
	In order to recognize potential information sources and factual evidence for each hallucination fragment $s_{i}$, this work creates a query set for each fragment by integrating its textual content and contextual information:
	\begin{equation}
		\mathcal{Q}_{i} = \{q_{i1}, q_{i2}, ..., q_{ik}\}
	\end{equation}
	
	Afterwards, we use the retrieval method proposed earlier to process each query $q_{ij}$, executing retrieval to obtain the evidence set as the final attribution result $\mathcal{E}_{i}$.
	
	Ultimately, the traceability result for each illusion fragment is represented as a quadruple:
	\begin{equation}
		\mathcal{T}_i = (s_i, \hat{y}_i^h, \hat{y}_i^e, \mathcal{E}_i)
	\end{equation}
	corresponding to hallucination text, hallucination type, error mechanism, and supporting evidence set.
	
\section{Experiments}

	\subsection{Experimental Setup}
	
	\subsubsection{Dataset Split and Evaluation Protocol}
	In the constructed hallucination tracing dataset, $70\%$ of the data is randomly allocated for model training, $10\%$ for parameter tuning and model selection, and the remaining $20\%$ is reserved as an independent test set for performance evaluation.	
	During evaluation, each hallucination segment $s_{i}$ serves as an individual query unit. 
	The system is required to retrieve and rank candidate evidence from the global evidence corpus $\mathcal{B}$. A model output is considered a successful attribution only when the returned results include evidence texts that form an equivalence class with the manually annotated evidence in the semantic space. This mechanism, by avoiding reliance on exact string matches, better reflects the practical requirements of real-world retrieval and fact verification scenarios.
	
	\subsubsection{Baselines}
	To comprehensively evaluate the effectiveness of our method in the hallucination attribution task, we opted for multiple representative sparse retrieval, dense retrieval, and semantic matching models as comparative baselines:
	
	\textbullet \ \ \textit{BM25}: A traditional sparse retrieval method based on term matching, serving as a lower-bound baseline without semantic modeling capabilities.
	
	\textbullet \ \ \textit{DPR}: A dense vector retrieval model with a dual-tower architecture, using pre-trained encoders to perform semantic matching between query and evidence embeddings.

	\textbullet \ \ \textit{Sentence-BERT}: Maps hallucinated fragments and evidence into a shared embedding space using a shared encoder, then performs nearest neighbor retrieval based on cosine similarity.
	
	\textbullet \ \ \textit{Cross Encoder Only}: Does not rely on vector indexes, instead, it performs semantic matching and full-text comparison directly on query-document pairs within the candidate set.
	
	\subsubsection{Implementation}
	During the semantic encoding phase, this paper employs the \textit{All-mpnet-base-v2} as the base encoding model. All hallucinated fragments and evidence texts are mapped into a high-dimensional vector space, with embedded vectors undergoing $L_{2}$ normalization. Vector indexes are constructed using \textit{FAISS}, employing inner-product similarity as the proximity search metric to achieve efficient retrieval across a large-scale evidence corpus (34,943 documents).
	During the reordering phase, the Transformer-based \textit{ms-marco-MiniLM-L-6-v2} model is introduced. It jointly encodes queries and candidate evidence, outputting refined semantic matching scores. This approach enhances semantic alignment capability while maintaining computational efficiency.
	
	All experiments were conducted on a single computer equipped with an NVIDIA RTX 5080 16G GPU and an AMD Ryzen 9950X processor.

	\subsection{Experimental Results}
	
	We constructed a hallucination attribution dataset based on outputs from the \textit{Qwen2.5-7B-Instruct} and \textit{Mistral-Small-24B-Instruct} models, and performed statistical analysis on hallucinations within the generated text. The results are summarized in Table~\ref{table:distribution}:

	\begin{table}[h]
		\centering
		\caption{Statistical Analysis of Hallucination Types and Error Mechanism Proportions in Qwen and Mistral Models}
		\small
		\begin{tabular}{llrr}
			\toprule
			\textbf{Category} & \textbf{SubType} & \textbf{Qwen($\%$)} &
			\textbf{Mistral($$\%)}\\
			\midrule
			
			\multirow{4}{*}{Hallucination Type}
			& Logic & 0.0033 & 0.0087 \\
			& Fabricate & 0.1028 & 0.0605 \\
			& Entity & 0.1676 & 0.0999 \\
			& Fact & \textbf{0.7263} & \textbf{0.8309} \\
			\midrule
			
			\multirow{5}{*}{Error Mechanism}
			& Context Leakage & 0.0054 & 0.0076 \\
			& Reasoning Failure & 0.0172 & 0.0168 \\
			& Overgeneralization & 0.0261 & 0.0244  \\
			& Entity Mismatch & 0.1228 & 0.1086 \\
			& Fabrication Heuristic & \textbf{0.8426} & \textbf{0.8436} \\
			
			\bottomrule
		\end{tabular}
		\label{table:distribution}
	\end{table}
	
	The outcomes indicate that the model exhibits a pronounced bias toward generating hallucinations, with factual errors being relatively common while logical errors are comparatively rare. Simultaneously, the model tends to employ a strategy of fabricating content, providing crucial evidence for subsequent hallucination attribution and error mechanism analysis.
	
	Table~\ref{table:ablation} presents ablation results for the evidence retrieval stage of our method. It can be observed that when relying solely on dense vector embeddings for retrieval, overall metric performance is constrained. Following the introduction of the \textit{Cross-Encoder} reordering module, the model achieves stable improvements in both ranking quality and retrieval precision. When further combined with the \textit{Multi-Query} strategy, recall performance and robustness are significantly enhanced. The complete method achieves optimal results across all evaluation metrics, demonstrating the effectiveness of the proposed retrieval framework in semantic alignment and evidence selection.
	
	\begin{figure}[h]
		\centering
		\begin{subfigure}[b]{0.48\linewidth}
			\centering
			\includegraphics[width=\linewidth]{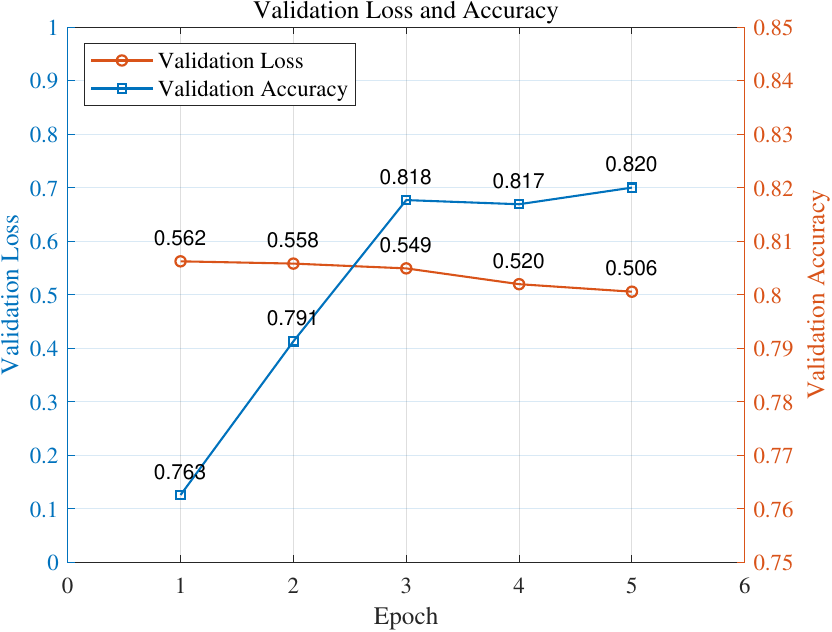}
			\caption{Hallucination Type Classifier}
			\label{fig:trace_auc_loss}
		\end{subfigure}
		\hfill
		\begin{subfigure}[b]{0.48\linewidth}
			\centering
			\includegraphics[width=\linewidth]{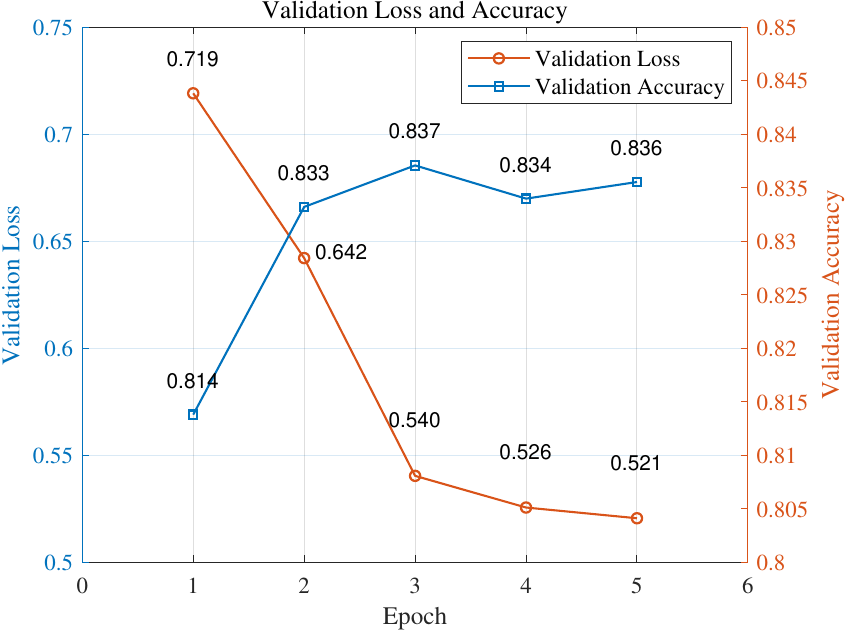}
			\caption{Error Mechanism Classifier}
			\label{fig:error_auc_loss}
		\end{subfigure}
		\caption{Loss and Accuracy During Classifier Training}
		\label{fig:classifier_loss_acc}
	\end{figure}
	
	\begin{table*}[h]
		\centering
		\caption{Hallucination Tracing Performance On Qwen And Mistral Dataset: HART vs. Baselines}
		\label{table:trace}
		\large
		\begin{tabular}{lllrrrr}
			\toprule
			\textbf{Dataset} & 
			\textbf{$k$} & 
			\textbf{Method} & 
			\textbf{Recall@$k$} & 
			\textbf{nDCG@$k$} & 
			\textbf{Joint SR@$k$} &
			\textbf{MRR @$k$} \\
			\midrule
			
			\multirow{15}{*}{Qwen}
			& \multirow{5}{*}{$k=1$}
			& BM25 & 0.1074 & 0.2659 & 0.0032 & 0.3016 \\
			& & DPR & 0.0349 & 0.0869 & 0.0014 & 0.1353 \\
			& & Sentence-BERT & 0.0859 & 0.2164 & 0.0020 & 0.1642\\
			& & Cross-Encoder & 0.0906 & 0.2269 & 0.0025 & 0.3298 \\
			& & Ours (HART) & \textbf{0.8024} & \textbf{0.7915} & \textbf{0.6265} & \textbf{0.8024}\\
			\cmidrule(lr){2-7}
			
			& \multirow{5}{*}{$k=2$}
			& BM25 & 0.2196 & 0.3896 & 0.0043 & 0.3612\\
			& & DPR & 0.0652 & 0.1290 & 0.0042 & 0.1635\\
			& & Sentence-BERT & 0.1592 & 0.2929 & 0.0223 & 0.2138 \\
			& & Cross-Encoder & 0.2046 & 0.3672 & 0.0238 & 0.3850 \\
			& & Ours (HART) & \textbf{0.8260} & \textbf{0.8173} & \textbf{0.6434} & \textbf{0.8142} \\
			\cmidrule(lr){2-7}
			
			& \multirow{5}{*}{$k=5$}
			& BM25 & 0.3945 & 0.4557 & 0.1899 & 0.3979 \\
			& & DPR & 0.1232 & 0.1686 & 0.0269 & 0.1841 \\
			& & Sentence-BERT & 0.3014 & 0.3407 & 0.1586 & 0.2487 \\
			& & Cross-Encoder & 0.3910 & 0.4208 & 0.2271 & 0.4135 \\
			& & Ours (HART) & \textbf{0.8331} & \textbf{0.8208} & \textbf{0.6493} & \textbf{0.8165} \\
			\midrule
			
			\multirow{15}{*}{Mistral}
			& \multirow{5}{*}{$k=1$}
			& BM25 & 0.0958 & 0.2888 & 0.0025 & 0.2882 \\
			& & DPR & 0.0330 & 0.0958 & 0.0011 & 0.1297\\
			& & Sentence-BERT & 0.0811 & 0.2402 & 0.0017 & 0.1379 \\
			& & Cross-Encoder & 0.1042 & 0.3085 & 0.0031 & 0.2663 \\
			& & Ours (HART) & \textbf{0.7522} & \textbf{0.7413} & \textbf{0.5352} & \textbf{0.7522} \\
			\cmidrule(lr){2-7}
			
			& \multirow{5}{*}{$k=2$}
			& BM25 & 0.2158 & 0.4259 & 0.0118 & 0.3382 \\
			& & DPR & 0.0607 & 0.1368 & 0.0020 & 0.1530 \\
			& & Sentence-BERT & 0.1591 & 0.3252 & 0.0042 & 0.1794 \\
			& & Cross-Encoder & 0.2151 & 0.4388 & 0.0076 & 0.3214 \\
			& & Ours (HART) & \textbf{0.7780} & \textbf{0.7685} & \textbf{0.5823} & \textbf{0.7651} \\
			\cmidrule(lr){2-7}
			
			& \multirow{5}{*}{$k=5$}
			& BM25 & 0.4345 & 0.5014 & 0.2161 & 0.3730 \\
			& & DPR & 0.1128 & 0.1749 & 0.0169 & 0.1699 \\
			& & Sentence-BERT & 0.3041 & 0.3698 & 0.1377 & 0.2144 \\
			& & Cross-Encoder & 0.4136 & 0.4833 & 0.2251 & 0.3515 \\
			& & Ours (HART) & \textbf{0.7895} & \textbf{0.7739} & \textbf{0.5921} & \textbf{0.7686} \\
			
			\bottomrule
		\end{tabular}
	\end{table*}

	\figurename~\ref{fig:classifier_loss_acc} displays the loss and accuracy curves of the hallucination type and error mechanism classifiers during training. On the validation set, the hallucination type classifier achieved an accuracy of $79.13\%$ , while the error mechanism classifier achieved $83.32\%$ accuracy. This demonstrates that the constructed classification model can reliably characterize the semantic category attributes and underlying error generation mechanisms of hallucination segments, providing structured prior information for subsequent evidence retrieval and attribution.
	
	Table~\ref{table:trace} contrasts the performance of the \textit{HART} method and baseline methods on the hallucination tracing dataset at different retrieval levels. Experimental results demonstrate that \textit{HART} significantly outperforms baseline methods in retrieval coverage, ranking quality, and end-to-end attribution success rate.
	Under the $k=1$ scenario, \textit{HART} exhibits strong prioritization capabilities, achieving Recall@1 scores of 0.8024 and 0.7522 on the \textit{Qwen} and \textit{Mistral} datasets respectively. This indicates that high-confidence external factual evidence is consistently placed at the top of retrieval results. As $k$ increases to 2 and 5, metric improvements are prioritized, suggesting that key evidence is concentrated in the top positions, demonstrating good ranking consistency and stability.
	
	Furthermore, the Joint SR@k metric for \textit{HART} significantly surpassed baseline methods across both datasets, illustrating that this framework enables effective hallucination attribution and external evidence retrieval. This achieves more reliable and interpretable end-to-end hallucination tracing capabilities.
\vspace{-1em}
\section{Conclusion}

	This paper pioneers a novel approach by modeling hallucination issues as real-world attribution and evidence tracing tasks from an external factual evidence attribution perspective. It introduces the unified \textit{HART} framework to establish a complete attribution chain from hallucinated fragments to factual grounds. From a broader perspective, this work advances hallucination research from a "detection-oriented" paradigm focused on error identification toward a real-world interpretability and attribution paradigm. Future research will explore cross-modal attribution and multi-hop evidence causal chain modeling to enhance the trustworthiness and explainability of large language models in high-risk application scenarios.

\bibliographystyle{ACM-Reference-Format}
\bibliography{sample-base}

\appendix
\section{Rationale Analysis of Evaluation Metrics}
\label{appendix_a}
	\subsection{Event Definition}
		Let the set of hallucination fragments in the dataset be:
		\begin{equation}
			\mathcal{H}_{true} = \{h_{1}, h_{2}, ..., h_{N}\}
		\end{equation}
		
		The set of genuine evidence corresponding to each hallucination fragment $h_{i}$ is defined as $\mathcal{E}(h_{i}) = \{e_{1}, e_{2}, \ldots, e_{G_{i}}\}$, where $G_{i}$ denotes the number of genuine evidence items for that fragment.
		
		For each search result $e_{j} \in \text{Top-k}$, define a Bernoulli random variable:
		\begin{equation}
			X_{j} = 
			\begin{cases}
				1, \ \ if \ e_{j} \in \mathcal{E}_{true} (h_{i}) \\
				0, \ \ else
			\end{cases}
		\end{equation}
		
		The event of successfully tracing the illusion fragment $h_{i}$ within the $\text{Top-}k$ can be expressed as:
		\begin{equation}
			Z_{k}(h_{i}) = \mathbb{I}(\bigcup_{j=1}^{k}\{X_{j} = 1\})
		\end{equation}
		
	\subsection{Expected Hit Probability}
		Assume the probability that each search result matches genuine evidence is $p_{j} = P(X_{j} = 1)$. Assuming the $\text{Top-}k$ results are mutually independent, the probability that $h_{i}$ successfully matches within the $\text{Top-}k$ is:
		\begin{equation}
			P(Z_{k}(h_{i}) = 1) = 1 - P(\bigcap_{j=1}^{k}\{X_{j} = 0\}) = 1 - \prod_{j=1}^{k}(1-p_{j})
		\end{equation}
		
		Further presumed that for all $j$, $p_{j} = p$, the above expression simplifies to:
		\begin{equation}
			P(Z_{k}(h_{i}) =  1) = 1 - (1 - p)^{k}
		\end{equation}
		
		Here, $p$ denotes the average probability that a single retrieval result successfully matches genuine evidence.
		
	\subsection{Recall@$k$ Expected Value}
		Define Recall@$k$ as the average recall rate over the entire dataset $\mathcal{H}$:
		\begin{equation}
			\text{Recall@}k = \frac{1}{|\mathcal{H}|} \sum_{i=1}^{N} Z_{k}(h_{i})
		\end{equation}
		
		Therefore, the expected value is:
		\begin{equation}
			\mathbb{E}[\text{Recall@}k] = \frac{1}{N} \sum_{i=1}^{N} \mathbb{E}[Z_{k}(h_{i})] = \frac{1}{N} \sum_{i=1}^{N} P(Z_{k}(h_{i}) = 1)
		\end{equation}
		
	\subsection{Equivalence \& Multi-Evidence}
		In actual datasets, each hallucination fragment may correspond to multiple equivalent or interchangeable pieces of evidence. Thus, the probability of a single retrieval interface hitting any genuine evidence becomes:
		\begin{equation}
			p=P(X_{j} = 1) = \sum_{g=1}^{G_{i}} p_{g}
		\end{equation}
		
		Where $p_{g}$ is the probability of retrieving the $g$th true evidence, the expected value of $\text{Recall@k}$ then becomes:
		\begin{equation}
			\mathbb{\text{Recall@$k$}} = 1 - \prod_{j=1}^{k}(1-p_{j}) \approx 1 - (1-\sum_{g=1}^{G_{i}})p_{g}
		\end{equation}

		Therefore, $\text{Recall@}k$ can be regarded as an empirical estimate of event probability. This expected value reflects the probability that the system successfully hits genuine evidence within a finite set of $\text{Top-}k$ retrieval results.

\section{Code Availability}
	To ensure reproducibility and facilitate further research, we provide an anonymized implementation of our proposed framework.
	The code is publicly available at:
	\begin{center}
		\url{https://anonymous.4open.science/r/KDD-AD53}
	\end{center}
	The repository contains the core implementation, evaluation pipeline, and instructions for reproducing the experimental results reported in this paper.
	All identifying information has been removed to comply with the single-blind review policy.

\end{document}